\documentclass[10pt,twocolumn,letterpaper]{article}

\usepackage{wacv}
\usepackage{times}
\usepackage{epsfig}
\usepackage{graphicx}
\usepackage{amsmath}
\usepackage{amssymb}

\usepackage{graphicx}
\usepackage{amsmath,amssymb} % define this before the line numbering.
\usepackage{color}
\usepackage{hyperref}       % hyperlinks
\usepackage{url}            % simple URL typesetting
\usepackage{booktabs}       % professional-quality tables
\usepackage{amsfonts}       % blackboard math symbols
\usepackage{nicefrac}       % compact symbols for 1/2, etc.
\usepackage{microtype}      % microtypography
\hypersetup{colorlinks=true, linkcolor=red}

\usepackage{soul}
\usepackage{multirow}

\usepackage{subcaption}     % for subfigures
\usepackage[utf8]{inputenc} % allow utf-8 input
\usepackage[T1]{fontenc}    % use 8-bit T1 fonts
\usepackage{hyperref}       % hyperlinks
\usepackage{url}            % simple URL typesetting
\usepackage{booktabs}       % professional-quality tables
\usepackage{amsfonts}       % blackboard math symbols
\usepackage{nicefrac}       % compact symbols for 1/2, etc.
\usepackage{microtype}      % microtypography
\usepackage{epsfig}
\usepackage{graphicx}
\hypersetup{colorlinks=true, linkcolor=red}

\usepackage{wrapfig}
\usepackage{algorithm}
\usepackage{algorithmic}
\usepackage{tikz}
\usepackage{pgfplots}
\usepackage{amsmath}
\usepackage{sci}
\usepackage{xfrac}
\usepackage{pgfplotstable}
\usepackage{filecontents}
\usepackage{adjustbox}
\usepackage{amssymb}
\usepackage{pifont}
\newcommand{\cmark}{\ding{51}}%
\newcommand{\xmark}{\ding{55}}%

\usepackage[normalem]{ulem}

\usetikzlibrary{arrows,calc,shapes,snakes,positioning,matrix,arrows,decorations.pathmorphing,decorations.text}
\usepackage[sort&compress,square,comma,numbers]{natbib}
\makeatletter
\renewcommand\bibsection%
{
  \section*{\refname
    \@mkboth{\MakeUppercase{\refname}}{\MakeUppercase{\refname}}}
}
\makeatother

\wacvfinalcopy % *** Uncomment this line for the final submission

\def\wacvPaperID{98} % *** Enter the wacv Paper ID here

\ifwacvfinal\fi
\begin{document}

%%%%%%%%% TITLE
%\title{TP-RNN: Triangular-Prism RNN for Action-Agnostic Human Pose Forecasting}
\title{Action-Agnostic Human Pose Forecasting}

% Authors at the same institution
%\author{First Author \hspace{2cm} Second Author \\
% Authors at different institutions
\author{Hsu-kuang Chiu, Ehsan Adeli, Borui Wang, De-An Huang, Juan Carlos Niebles\\
Stanford University\\
{\tt\small \{hkchiu, eadeli, wbr, dahuang, jniebles\}@cs.stanford.edu}
% \and
% Second Author \\
}

\maketitle
\ifwacvfinal\thispagestyle{empty}\fi

%%%%%%%%% ABSTRACT
\begin{abstract}
Predicting and forecasting human dynamics is a very interesting but challenging task with several prospective applications in robotics, health-care, {\it etc}. Recently, several methods have been developed for human pose forecasting; however, they often introduce a number of limitations in their settings. For instance, previous work either focused only on short-term or long-term predictions, while sacrificing one or the other. Furthermore, they included the activity labels as part of the training process, and require them at testing time. %(\ie, either building models for each single activity separately or incorporating ground-truth activity labels as input features to their models). 
These limitations confine the usage of pose forecasting models for real-world applications, as often there are no activity-related annotations for testing scenarios. In this paper, we propose a new \textit{action-agnostic} method for short- and long-term human pose forecasting. %(\ie, training the model regardless of the activity labels). 
To this end, we propose a new recurrent neural network for modeling the hierarchical and multi-scale characteristics of the human dynamics, denoted by triangular-prism RNN (TP-RNN). Our model captures the latent hierarchical structure embedded in temporal human pose sequences by encoding the temporal dependencies with different time-scales. For evaluation, we run an extensive set of experiments on \textit{Human 3.6M} and \textit{Penn Action} datasets and show that our method outperforms baseline and state-of-the-art methods quantitatively and qualitatively. Codes are available at \url{https://github.com/eddyhkchiu/pose_forecast_wacv/}
\end{abstract}

%%%%%%%%% BODY TEXT
\section{Introduction}

Humans are able to %One of the \fix{fundamental human abilities} \jc{(JC: do we have a source to say it's fundamental?)} is to 
predict how their surrounding environment may change and how other people move. This inclination and aptitude is crucial to make social life and interaction with others attainable \cite{argyle1972non}. As such, to create machines that can interact with humans seamlessly, it is very important to convey the ability of predicting short- and long-term future of human dynamics based on the immediate present and past. Recently, computer vision researchers attempted predicting human dynamics from images \cite{chao2017forecasting}, or through time in videos \cite{vondrick2016generating,barsoum2017hp,schydlo2018anticipation}. Human dynamics are mainly defined as a set of structured body joints, known as poses \cite{sigal2014human}. Predicting human dynamics is hence delineated by predicting the course of changes in human poses \cite{wei2016convolutional,jain2016structural,martinez2017simple,chao2017forecasting,walker2017pose,toyer2017human}.

% \begin{figure}[t!]
%     \centering
%     \begin{tikzpicture}
%         \node[inner sep=0pt] (figure1) at (0.7,0)
%         {\includegraphics[width=0.9\linewidth]{Img/walking1_final_part_merge_downsampled.png}};
%         \draw [<-,>=stealth,thick] (-4.7,2.6) -- (-1.7,2.6) node[midway,fill=white] {Past};
%         \draw [->,>=stealth,thick] (-1.5,2.6) -- (6.2,2.6) node[midway,fill=white] {Future};
%         \node[inner sep=0pt,text width=1cm,align=center] (gt) at (-5.4,1.8) {\tiny Ground Truth};
%         \node[inner sep=0pt] (t1) at (-5.4,0.6) {\tiny TP-RNN};
%         \node[inner sep=0pt] (t2) at (-5.4,-0.5) {\tiny Right Arm};
%         \node[inner sep=0pt] (t3) at (-5.4,-1.3) {\tiny Left Leg};
%         \node[inner sep=0pt] (t4) at (-5.4,-2.1) {\tiny Torso};
%     \end{tikzpicture}
%     \caption{Illustration of the ground truth ($1^\text{st}$ row), the forecasted poses by our method ($2^\text{th}$ row), and individual body parts (next 3 rows). The movement patterns of body parts in time reveals that different part movements depend on each other, and each part has its own temporal scale. Hence, hierarchical multi-scale modeling can encode the latent structures of human dynamics.}
%     \label{fig:fig1}
% \end{figure}
\begin{figure}[t!]
    \centering
    \begin{tikzpicture}
        \node[inner sep=0pt] (g1)         {\includegraphics[width=0.12\linewidth]{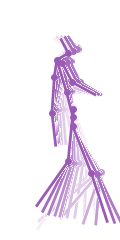}};
        \node[inner sep=0pt, right=0.0 of g1] (g2)         {\includegraphics[width=0.12\linewidth]{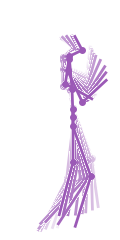}};
        \node[inner sep=0pt, right=0.0 of g2] (g3)         {\includegraphics[width=0.12\linewidth]{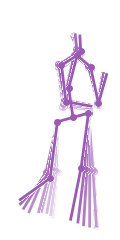}};
        \node[inner sep=0pt, right=0.0 of g3] (g4)         {\includegraphics[width=0.12\linewidth]{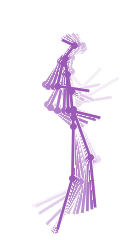}};
        \node[inner sep=0pt, right=0.0 of g4] (g5)         {\includegraphics[width=0.12\linewidth]{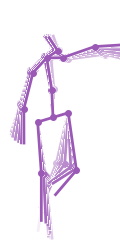}};
        \node[inner sep=0pt, right=0.0 of g5] (g6)         {\includegraphics[width=0.12\linewidth]{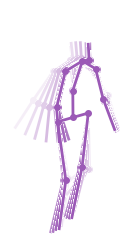}};
        \node[inner sep=0pt, right=0.0 of g6] (g7)         {\includegraphics[width=0.12\linewidth]{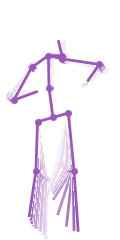}};
        \node[inner sep=0pt, right=0.0 of g7] (g8)         {\includegraphics[width=0.12\linewidth]{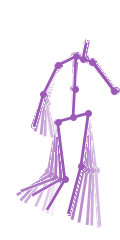}};
        
        \node[inner sep=0pt, below=-0.1cm of g1] (o1)         {\includegraphics[trim={0 0 0 1cm}, clip,  width=0.12\linewidth]{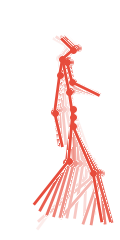}};
        \node[inner sep=0pt, below=0.0 of g2] (o2)         {\includegraphics[trim={0 0 0 1cm}, clip, width=0.12\linewidth]{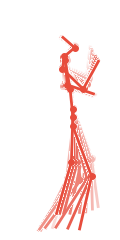}};
        \node[inner sep=0pt, below=0.0 of g3] (o3)         {\includegraphics[trim={0 0 0 1cm}, clip, width=0.12\linewidth]{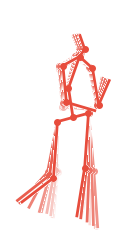}};
        \node[inner sep=0pt, below=0.0 of g4] (o4)         {\includegraphics[trim={0 0 0 1cm}, clip, width=0.12\linewidth]{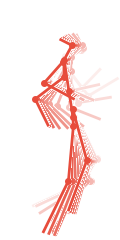}};
        \node[inner sep=0pt, below=0.0 of g5] (o5)         {\includegraphics[trim={0 0 0 1cm}, clip, width=0.12\linewidth]{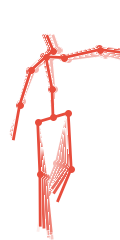}};
        \node[inner sep=0pt, below=0.0 of g6] (o6)         {\includegraphics[trim={0 0 0 1cm}, clip, width=0.12\linewidth]{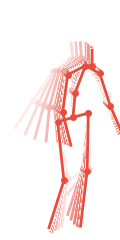}};
        \node[inner sep=0pt, below=0.0 of g7] (o7)         {\includegraphics[trim={0 0 0 1cm}, clip, width=0.12\linewidth]{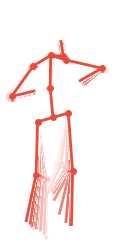}};
        \node[inner sep=0pt, below=0.0 of g8] (o8)         {\includegraphics[trim={0 0 0 1cm}, clip, width=0.12\linewidth]{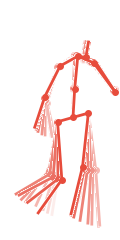}};

        \node[inner sep=0pt,above=-0.2cm of g1] (t1) {\tiny Walking};
        \node[inner sep=0pt,right=0.5cm of t1] (t2) {\tiny Eating};
        \node[inner sep=0pt,right=0.5cm of t2] (t3) {\tiny Smoking};
        \node[inner sep=0pt,right=0.3cm of t3] (t4) {\tiny Discussion};
        \node[inner sep=0pt,right=0.0cm of t4,text width=1cm,align=center] (t5) {\baselineskip=8pt\tiny Walking together\par};
        \node[inner sep=0pt,right=0.2cm of t5,text width=0.9cm,align=center] (t6) {\baselineskip=8pt\tiny Walking dog\par};
        \node[inner sep=0pt,right=0.1cm of t6] (t7) {\tiny Waiting};
        \node[inner sep=0pt,right=0.3cm of t7,text width=1cm,align=center] (t8) {\baselineskip=8pt\tiny Talking on the phone\par};
        
        \node[inner sep=0pt,left=0.0cm of g1, rotate=90, anchor=base] (t00) {\scriptsize Ground-truth};
        \node[inner sep=0pt,left=0.0cm of o1,rotate=90, anchor=base] (t01) {\scriptsize TP-RNN};

    \end{tikzpicture}
    \caption{Ground-truth pose sequences (first row) and forecasted ones by our method (second row). Solid colors indicate later time-steps and faded ones are older. The course of changes in the predicted and ground-truth poses resemble similar patterns. Furthermore, body-part movement patterns show that different parts depend on each other, but with varied temporal scales. Hence, hierarchical multi-scale modeling may encode the latent structures of human dynamics.}
    \label{fig:fig1}
\end{figure}
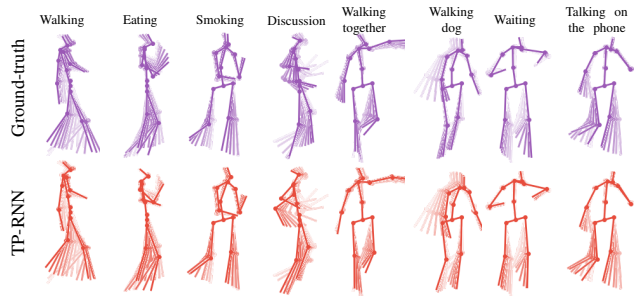

Detecting and predicting poses has long been an interesting topic in the computer vision community \cite{sigal2014human,wei2016convolutional,moeslund20003d}. Recently, several methods have been introduced for forecasting human poses in the near future \cite{jain2016structural,martinez2017simple,walker2017pose,toyer2017human,chao2017forecasting}. In a recent work, Martinez \etal~\cite{martinez2017human} noted that, although great advancements in pose forecasting has been achieved by prior works, they often fail to generate realistic human poses, especially in short-term predictions, and in most cases they fail to even outperform the zero-velocity predictor (\ie, repeating the very last seen pose as predictions for the future). Ghosh \etal~\cite{ghosh2017learning}, with reference to \cite{martinez2017human}, attributed this finding to the side-effects of curriculum learning (such as in \cite{chao2017forecasting}), commonly practiced %among researchers 
for temporal forecasting. With such observations, some previous works focused on short-term forecasting of human poses \cite{martinez2017human,jain2016structural}, and some others exclusively aimed attention at long-term predictions  \cite{ghosh2017learning,barsoum2017hp,toyer2017human}. However, most of the previous methods achieve reasonable performance by incorporating action labels as extra data annotation in their models, \ie, they either trained pose forecasters on each action class separately (\eg, \cite{jain2016structural,fragkiadaki2015recurrent}) or incorporated the action labels as an extra input to the model and concluded that including action labels improves the results (\eg, \cite{martinez2017human}). Although action labels are not hard to acquire for training samples, but the use of labels for testing videos as an input of the model is \textit{unrealistic} and makes the introduced models \textit{unusable} for real-world applications, as action labels are not available during testing \cite{mosabbeb2014multi}. Unlike these previous works, our method learns a pose forecaster regardless of their action class. We propose an action-agnostic model for pose forecasting by implicitly encoding the short- and long-term dependencies within actions.

In this paper, we propose a new recurrent neural network (RNN) model for forecasting human poses in both short- and long-term settings. To model human dynamics and to capture the latent hierarchical structure in the temporal pose sequences, we encode the temporal dependencies of different time-scales in a hierarchical interconnected sequence of RNN cells. Called \textit{Triangular-Prism Recurrent Neural Network (TP-RNN)}, our proposed method contains a new multi-phase hierarchical multi-scale RNN architecture that is tailored for modeling human dynamics in visual scenes. Different from the original hierarchical multi-scale RNNs (HM-RNN) for representation of natural language sequences \cite{chung2017hierarchical}, our architecture redefines hierarchies and multi-scale interconnections to accommodate human dynamics. Sequences of human poses through time involve hierarchical and multi-scale structures, as movements of different body-parts (and joints) depend on each other. Besides, each of these parts (and joints) have distinct motion patterns and hence different temporal scales for particular activities. For instance, during `walking', arms and legs move in a shorter temporal scale (\ie, more frequently) compared to the torso, which is potentially in a longer temporal scale (see Fig. \ref{fig:fig1}). Learning the hierarchical multi-scale dynamics of changes in human poses enables TP-RNN to construct an implicit encoding of short- and long-term dependencies within action classes, and hence be able to predict future sequences without the demand for the supervising signal from action labels. 

%HM-RNN \cite{chung2017hierarchical} learns the latent representation of natural language sequences in different hierarchies, such as words, phrases, and sentences to build character-level language models for predicting future sequences \cite{el1996hierarchical,bahdanau2016end}. Such a hierarchical and multi-scale architecture is able to more precisely capture temporal information at different hierarchies and scales for language models. We argue that multi-scale temporal information in different hierarchies are also beneficial in modeling human dynamics. Although not evident as the sequential language models, 

Our model takes pose velocities (\ie, differences between the current and the immediate previous poses, ignoring the division by the constant time duration between two consecutive frames) as inputs and outputs predictions in the same space of velocities. As opposed to the previous works \cite{fragkiadaki2015recurrent,jain2016structural,walker2017pose} that focused on predicting sequences of poses (as structured objects) in the forms of either joint angles (\eg, \cite{fragkiadaki2015recurrent}) or joint locations (such as in \cite{jain2016structural}), we argue that forecasting in the velocities space boosts prediction power since human poses change slightly in narrow time-steps. Different from \cite{martinez2017human}, in which residual connections were applied on top of RNN %the RNN sequence-to-sequence model 
(using residuals as outputs only, while inputs are poses), our method uses velocities as both inputs and outputs and shows significantly improved forecasting results.

To evaluate the proposed method, we run an extensive set of experiments on Human 3.6M \cite{ionescu2014human3} and Penn Action \cite{zhang2013from} datasets, and compare the results with several baseline and state-of-the-art algorithms on these datasets. The comparison shows that our method outperforms others in terms of the mean angle error (MAE) on Human 3.6M and the Percentage of Correct Keypoint (PCK) score on Penn Action. Our action-agnostic method leads to superior results in cases of both short- and long-term predictions (some are visualized in Fig. \ref{fig:fig1}) even in comparison to the methods designed specifically for short- or long-term predictions or methods that use action labels as inputs to their models.

In summary, the contributions of this paper are three-fold: (1) we propose an action-agnostic model that trains the pose forecaster regardless of action classes; (2) we propose a new model, TP-RNN, inspired by the hierarchical multi-scale RNN from NLP research, for forecasting human dynamics. TP-RNN implicitly encodes the action classes and, unlike previous methods, does not require external action labels during training; (3) we show that operating in the velocity space (\ie, using pose velocities as both inputs and outputs of the network) improves the results of our model.

% efficiently delivering long-term dependencies with fewer updates at the high-level layers, which mitigates the vanishing gradient problem, 

% hidden units to the higher layers that focus on modelling long-term dependencies and less hidden
% units to the lower layers which are in charge of learning short-term dependencies

\section{Related Works}
In this Section, we review the relevant literature on human motion, activity, and pose forecasting, along with the previous works on hierarchical and multi-scale RNNs (and Long Short-Term Memory cells, \ie, LSTMs). %used in vision or natural language contexts.

%\subsection{Predicting Motion and Human Dynamics}
\noindent\textbf{Predicting Motion and Human Dynamics:}
The majority of the recent works on motion representation has mainly focused on anticipating the future at the pixel level. For instance, generative adversarial networks (GANs) were used to generate video pixels \cite{vondrick2016generating,mathieu2016deep}, and RNNs for anticipating future video frames \cite{mahjourian2017geometry}. To predict dense trajectories, Walker \etal~\cite{walker2015dense} used a CNN, and others have used random forests \cite{pintea2014deja} or variational auto-encoders \cite{walker2016uncertain}. 
Other works targeted predicting the future in forms of semantic labels (\eg, \cite{lan2014hierarchical,luc2017predicting,walker2016uncertain}) or activity labels (\eg, \cite{walker2017pose,soomro2018online,aliakbarian2017encouraging,arzani2017structured}). Human dynamics, however, could be better characterized by 2D \cite{cao2017realtime,rogez2008randomized,rogez2017lcr} or 3D \cite{sandriluka2010monocular,ye2011accurate,tome2017lifting,mehta2017vnect} poses, and several works attempted to detect these poses from images or videos \cite{ye2011accurate,shotton2011real,tome2017lifting}. Modeling human motions is commonly defined in two different ways: \textit{probabilistic and state transition models} (such as Bayesian and Gaussian processes \cite{wang2008gaussian} or hidden Markov models \cite{wu2014leveraging}), and \textit{deep learning methods}, in particular RNNs and LSTMs, \eg, \cite{ghosh2017learning,martinez2017human,jain2016structural}. For instance, Jain \etal~\cite{jain2016structural} proposed a structural RNN to cast an arbitrary spatio-temporal graph as a RNN and use it for modeling human pose in temporal video sequences. %In general, deep learning models have shown better results in modeling temporal dynamics compared to traditional methods. 
In this work, we propose a new multi-phase hierarchical multi-scale RNN for modeling human dynamics to forecast poses. 

%\subsection{Human Pose Forecasting}
\noindent\textbf{Human Pose Forecasting:}
Forecasting human poses in images and video sequences is relatively new compared to predicting image or video pixels. Although it can be a very useful task with great applications (\eg, in predictive surveillance, patient monitoring, \etc), just recently researchers have paid more attention to it \cite{chao2017forecasting,martinez2017simple,martinez2017human,walker2017pose,jain2016structural,toyer2017human,barsoum2017hp}.

Specifically, Chao \etal~\cite{chao2017forecasting} proposed a 3D Pose Forecasting Network (3D-PFNet) for forecasting human dynamics from static images. Their method integrates recent advances on single-image human pose estimation and sequence prediction. %However, they do not incorporate the rich temporal and movement history information of the human subjects present in the image, and try to forecast the pose solely based on the image (\ie, single frame or snapshot of a video). 
In another work, \cite{martinez2017simple} introduced a method to predict 3D positions of the poses, given their 2D locations. Barsoum \etal~\cite{barsoum2017hp} proposed a sequence-to-sequence model for the task of probabilistic pose prediction, trained with an improved Wasserstein GAN \cite{arjovsky2017wasserstein}. Walker \etal~\cite{walker2017pose} proposed a method based on variational autoencoders and GANs to predict possible future human movements (\ie, poses) and then predict future frames. Fragkiadaki \etal~\cite{fragkiadaki2015recurrent} proposed two architectures for the task of pose prediction, one denoted by LSTM-3LR (3 layers of LSTM cells) and the second one as ERD (Encoder-Recurrent-Decoder). These two models are based on a sequence of LSTM units. %Similar to SRNN \cite{jain2016structural}, LSTM-3LR and ERD gradually add noise to the input during training, and conduct noise scheduling, to make the network robust to prediction errors. 
Martinez \etal~\cite{martinez2017human} used a variation of RNNs to model human motion with the goal of learning time-dependent representations for human motion prediction synthesis in a short-term. Three key modifications to recent RNN models were introduced, in the architecture, loss function, and the training procedures. In another work, B{\"u}tepage \etal~\cite{butepage2017deep} proposed an encoding-decoding network that learns to predict future 3D poses from the immediate past poses, and classify the pose sequences into action classes. These two methods
%, unlike other motion prediction models, 
incorporate a high-level supervision in the form of action labels, which itself improves the performance. However, in many real world applications of human motion analysis there are no motion or activity labels available during inference time.

%\subsection{Hierarchical Multi-Scale Recurrent Neural Network (HM-RNN)}
\noindent\textbf{Hierarchical Multi-Scale RNNs: }
%NEW
Our proposed architecture is inspired by the hierarchical multi-scale recurrent neural networks (HM-RNN) introduced in \cite{chung2017hierarchical}. HM-RNN builds on multi-scale RNNs \cite{koutnik2014clockwork} that model high-level abstraction changes slowly with temporal coherency while low-level abstraction has quickly changing features sensitive to the precise local timing \cite{el1996hierarchical}. %Multi-scale RNNs group hidden units into multiple modules of different time-scales. HM-RNN architecture introduced hierarchies, which delivers long-term dependencies with fewer updates at the high-level layers and mitigates the vanishing gradient problem. 
This architecture is able to learn the latent representation of natural language sequences in different hierarchies (\eg, words, phrases, and sentences) to build character-level language models for predicting future sequences \cite{el1996hierarchical,bahdanau2016end}. We observe that multi-scale temporal information at different hierarchical levels can be  beneficial in modeling human dynamics. However, it is difficult to adopt this approach directly because we do not have clear-cut temporal boundaries as in natural language data. %Although not evident as the sequential language models,  We further extend this model for visual data for representing human pose sequences, which also enjoy from hierarchical structures, 

\section{Triangular-Prism RNN (TP-RNN)}
% \begin{wrapfigure}{R}{0.49\textwidth}
%     \centering
%     \vspace{-20pt}
%     \includegraphics[width=\linewidth]{Img/tp-rnn.pdf}
%     \vspace{-10pt}
%     \caption{Illustration of TP-RNN with the scale of two and three levels of hierarchy. Model in each layer (same color) share learning weights. Links between layers are removed for clarity (see Fig. \ref{fig:arch}).}
%     \label{fig:prism}
%     \vspace{-15pt}
% \end{wrapfigure}
%\subsection{Architecture}
%In this section, we present our new neural architecture for human pose prediction called \textbf{Triangular-Prism Recurrent Neural Network}. 

\begin{figure*}[t!]
\centering
\includegraphics[width=0.9\linewidth]{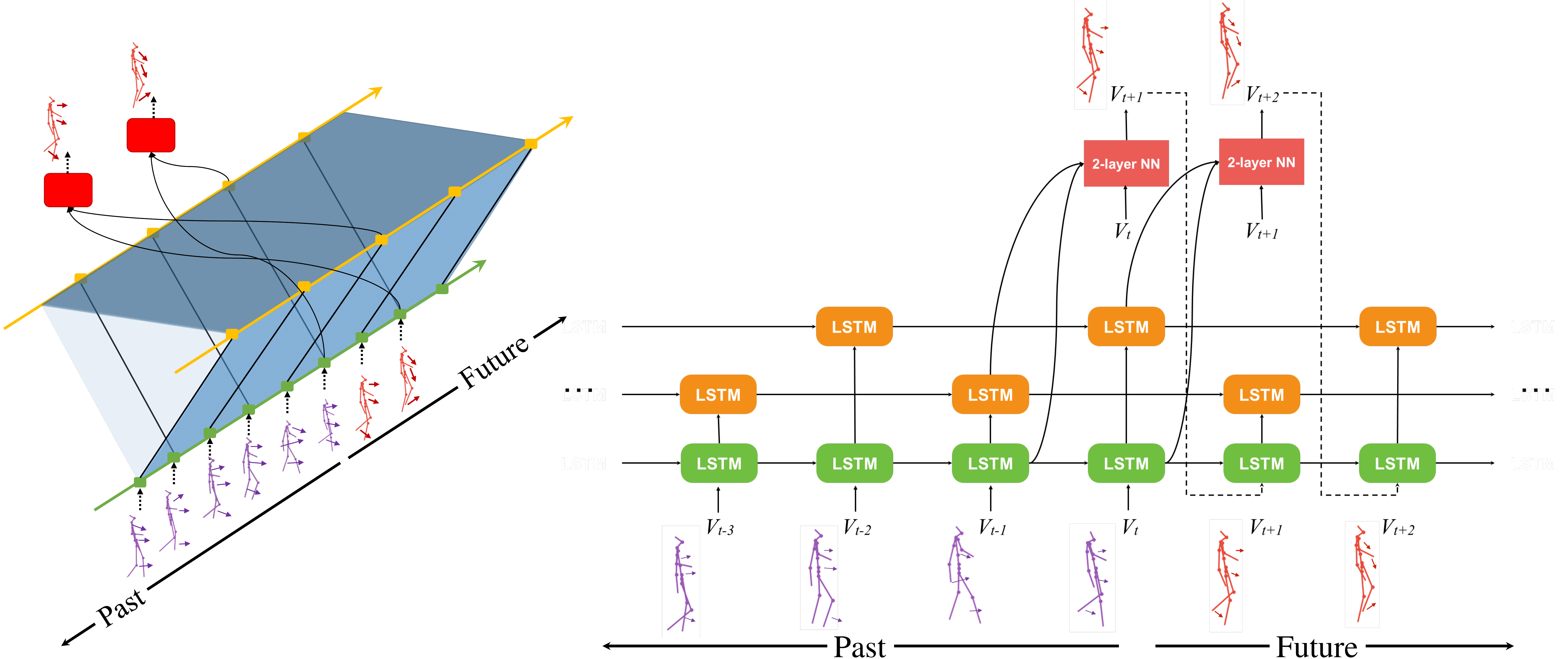}
\caption{Architecture of TP-RNN with $K = 2$ and $M = 2$. Left: 3D view of the triangular-prism RNN. Right: 2D projection.}
\label{fig:arch}
\end{figure*}

As discussed earlier, sequences of human poses can be subsumed under hierarchical and multi-scale structures, since movements of different body parts hinge on movements of other parts. Also, each part often has distinct motion patterns and hence different temporal scales when performing particular activities. Therefore, in contrast to the classical single-layer LSTM or RNN architectures (such as in \cite{jain2016structural}) or stacked LSTMs (\eg, in \cite{donahue2015long,wang2015long}), we introduce multi-phase hierarchical multi-scale upper layers of LSTM sequences to better learn the longer-term temporal relationships between different time-steps in a series of different granularities.

The inputs and outputs of the model, as mentioned earlier, are velocities. % (differences between poses in two consecutive frames, ignoring the division on the constant time duration). ALREADY DEFINED 
Let the pose in time $t$ be identified by $P_t$, then the velocity in time $t$ can be defined as $V_t = P_t - P_{t-1}$. Therefore, for any given granularity coefficient $K$ and the number of levels $M$, we define a multi-phase hierarchical multi-scale RNN with scale $K$ and $M$ levels. On the first level, we have a regular LSTM sequence taking the velocity information at each time-step as inputs. Then, on the second level, we define $K$ different sequences of LSTM units, with each sequence only taking the inputs from the LSTM units on the first level at time-steps that are congruent modulo $K$. For example, if $K = 2$, then we have two LSTM sequences at level 2, with the first one taking inputs from $t = \{1,3,5,\dots\}$ and the second one from $t = \{2,4,6,\ldots\}$ (see Fig. \ref{fig:arch} for illustrations). Note that these LSTMs in the same level of the hierarchy share weights and this shifting scheme is actually used as a data augmentation strategy to learn longer-term dependencies in a more reliable way. Similarly, if we define a third level in the hierarchy, for each of the $K$ LSTM sequences on the second level, we will have $K$ different LSTM sequences each taking congruent-modulo-$K$ inputs from it, resulting in a total of $K^2$ LSTM sequences in the third level. This process of spawning new higher-level LSTM sequences over the hierarchy continues for ${M-1}$ levels, which will have $K^{M-1}$ LSTM sequences in the $M$th level. Therefore, logically, we have a total of $(K^{M-1} + K^{M-2} + \ldots + K^3 + K^2 + K + 1)$ LSTM sequences in the whole architecture, while only $M$ different ones are kept physically (since the LSTM sequences in each level share weights). 
For the sampling stage, we introduce a two-layer fully-connected network to generate the velocity predictions given the velocity at the current time-step and the corresponding hidden units across all hierarchies. Fig. \ref{fig:arch} illustrates an example of our architecture for $K = M = 2$. 

Our hierarchical model is inspired by HM-RNN \cite{chung2017hierarchical}, however, as discussed earlier, we cannot directly apply HM-RNN to our task, due to the differences between human dynamics and natural language sequences. TP-RNN models short-term dependencies in lower levels of the hierarchy and long-term temporal dependencies in higher levels, and hence can capture the latent hierarchical structure in different time-scales. Different from HM-RNN, since human dynamics (unlike language models) do not have natural boundaries in the sequences, TP-RNN uses the outputs of all hierarchy levels to represent the sequence and predict the next element in the sequence. Moreover, instead of having only one RNN layer in each level of the hierarchies, as we move up in the hierarchy, TP-RNN decreases the resolution by one unit but has multiple RNN layers in the higher levels, to capture the temporal dynamics from different phases of its lower hierarchy level. All RNN layers in each single hierarchy level share parameters; although this scheme does not increase model parameters, shifting phases from each lower level to create their immediate higher level RNNs helps in augmenting the data and learning better models at each level. Therefore, as we go up in the hierarchy, more parallel RNN layers are incorporated, and hence we chose the name triangular-prism RNN (see Fig. \ref{fig:arch}). This architecture design provides the following advantages over HM-RNN for modeling human dynamics: (1) the lowest layer RNN can learn the finest grained scale motion dynamics without the interference from the higher levels, and higher levels capture different characteristics of the dynamics each at a certain scale; (2) during the prediction of each time-step, we have the most up-to-date RNN outputs from different hierarchies, each of which carries temporal motion information with different scales. On the contrary, HM-RNN does not provide the most up-to-date larger scale temporal information when the current prediction time-step is not right after a boundary.

\section{Experiments}
We evaluate our method on two challenging datasets. The results are analyzed and compared with baseline and state-of-the-art techniques, both quantitatively and qualitatively.

In our architecture, we use LSTMs with hidden size 1024 as the RNN cells (the orange and the green blocks in Fig. \ref{fig:arch}). For the final pose velocity generation networks (the red blocks in Fig. \ref{fig:arch}), we use 2 fully-connected layers with hidden sizes 256 and 128, followed by a Leaky-ReLU non-linearity layer. The training setup is similar to \cite{martinez2017human}. The optimization uses mini-batch stochastic gradient descent with batch size 16, clipping the gradients up to $\ell_2$-norm value of 5. The learning rate is initialized to 0.01 and decayed along the training iterations. We train for 100,000 iterations and record the performance coverage in the end.

\subsection{Datasets}
To test the performance of our model for human pose forecasting, we run extensive experiments using Human 3.6M \cite{ionescu2014human3} and Penn Action \cite{zhang2013from} datasets. 

%\subsubsection{The Human 3.6M dataset}
\noindent\textbf{Human 3.6M dataset:}
The Human 3.6M dataset \cite{ionescu2014human3} is one of the largest publicly available datasets of human motion capture data. %, used by a series of previous works on human motion modeling and prediction, thus making it easy to directly and fairly compare the performance of our new method with other state-of-the-art benchmark methods. 
This dataset contains video sequences of a total of 15 different human activity categories, %such as `walking', `sitting', `taking photos', and `walking dogs'. Each activity is 
each performed by seven actors in two different trials. The videos %in the Human 3.6M dataset 
were recorded at 50Hz (\ie, 20ms between each two consecutive pose frames). Following previous work \cite{jain2016structural,martinez2017human}, in our experiments, we downsample the pose sequence by 2. In the dataset, each pose is represented as exponential map representations of 32 human joints in the 3D space, and during evaluation, %as in previous works, 
we employ the measurement of the Euclidean distance between the ground-truth pose and our predicted pose in the angle space as the error metric. Consistent with the previous work, we also use Subject 5 as the test data and Subjects 1, 6, 7, 8, 9, 11 as training. Similar to \cite{jain2016structural,martinez2017human}, we train our models using the past 50 frames (2000ms) as the input sequence, and forecast the future 25 frames (1000ms). The training loss is calculated by the mean angle error (MAE) from each of the predicted future frames. %, similar to the setting of long-term prediction from \cite{jain2016structural,martinez2017human}. 
%Besides, we train our model in an action-agnostic setup, using all the 15 action categories of Human 3.6M dataset, instead of training individual models for each action class. %by only using sequence data from individual actions.

%\subsubsection{The Penn Action dataset}
\noindent\textbf{Penn Action dataset:}
The second dataset we experiment on is the Penn Action dataset \cite{zhang2013from}, which contains 2326 video sequences of 15 different actions and human joint annotations for each sequence. Each human pose is represented by 13 human joint coordinates in the 2D space. Following the same data split of  \cite{zhang2013from,chao2017forecasting}, 1258 video sequences are used for training and the remaining 1068 video sequences are used for testing. For this dataset, the previous state-of-the-art, 3D-PFNet \cite{chao2017forecasting}, takes the first frame image as input, and outputs the poses extracted from that frame and the future 15 frames, resulting in total of 16 frame poses. The model performance is evaluated using PCK@0.05 \cite{chao2017forecasting}. For the experiments on this dataset, we use a single pose in a past frame (ignoring the frame image) as the input, and the outputs are the predictions of poses of the future 16 frames. Although the input format of \cite{chao2017forecasting} is slightly different from ours, it is still a fair comparison of pose forecasting capabilities.

% \subsection{Setup}
% In our architecture, we use LSTMs with hidden size 1024 as the RNN cells (the orange and the green blocks in Fig. \ref{fig:arch}). For the final pose velocity generation networks (the red blocks in Fig. \ref{fig:arch}), we use 2 fully-connected layers with hidden sizes 256 and 128, followed by a Leaky-ReLU non-linearity layer. The training setup is similar to \cite{martinez2017human}. The optimization uses mini-batch stochastic gradient descent with batch size 16, clipping the gradients up to $\ell_2$-norm value of 5. The learning rate is initialized to 0.01 and decayed along the training iterations. We train for 100,000 iterations and record the performance coverage in the end.

\begin{table*}[t!]
\caption{MAE for four action classes %`walking', `eating', `smoking', and `discussion' 
in the short-term forecasting experiment (prior-work results from \cite{martinez2017human}). In each column, the best obtained results are typeset in boldface and the second best are underlined. AA: Action-Agnostic, N/A: Not Applicable.}
\vspace{-15pt}
\setlength{\tabcolsep}{4pt}
\label{tab:res1}
\begin{center}
{\small
\begin{tabular}{ l|c|cccc|cccc|cccc|cccc  }
 \hline
 & AA & \multicolumn{4}{c|}{Walking} & \multicolumn{4}{c|}{Eating} & \multicolumn{4}{c|}{Smoking} & \multicolumn{4}{c}{Discussion} \\
 \hline
 milliseconds               & &   80 &  160 &  320 &  400 &   80 &  160 &  320 &  400 &   80 &  160 &  320 &  400 &   80 &  160 &  320 &  400 \\
 \hline
 ERD \cite{fragkiadaki2015recurrent}        & \xmark    & 0.93 & 1.18 & 1.59 & 1.78 & 1.27 & 1.45 & 1.66 & 1.80 & 1.66 & 1.95 & 2.35 & 2.42 & 2.27 & 2.47 & 2.68 & 2.76\\ 
 LSTM-3LR \cite{fragkiadaki2015recurrent}   &  \xmark    & 0.77 & 1.00 & 1.29 & 1.47 & 0.89 & 1.09 & 1.35 & 1.46 & 1.34 & 1.65 & 2.04 & 2.16 & 1.88 & 2.12 & 2.25 & 2.23\\ 
 SRNN \cite{jain2016structural}             &  \xmark    & 0.81 & 0.94 & 1.16 & 1.30 & 0.97 & 1.14 & 1.35 & 1.46 & 1.45 & 1.68 & 1.94 & 2.08 & 1.22 & 1.49 & 1.83 & 1.93\\
 Residual \cite{martinez2017human}          &  \xmark    & \underline{0.28} & \underline{0.49} & \underline{0.72} & \underline{0.81} & \underline{0.23} & \underline{0.39} & \underline{0.62} & \underline{0.76} & \underline{0.33} & 0.61 & 1.05 & 1.15 & \underline{0.31} & 0.68 & 1.01 & \underline{1.09} \\
% Residual \cite{martinez2017human}          &  \cmark    &  \\
 Zero-velocity                              & N/A & 0.39 & 0.68 & 0.99 & 1.15 & 0.27 & 0.48 & 0.73 & 0.86  & \textbf{0.26} & \underline{0.48} & \underline{0.97} & \underline{0.95} & \underline{0.31} & \underline{0.67} & \textbf{0.94} & \textbf{1.04} \\
%  \hline
%  Single Layer (Pose)                    & 0.42 & 0.70 & 0.93 & 0.98 & 0.32 & 0.52 & 0.74 & 0.88 \\
%  Single Layer (Vel.)                    & \textbf{0.25} & \underline{0.42} & \underline{0.60} & \underline{0.67} & \textbf{0.20} & \underline{0.33} & \underline{0.54} & \textbf{0.67} \\
%  \hline
%  Stacked 2-Layer (Vel.)                 & \textbf{0.25} & \textbf{0.41} & \underline{0.60} & \underline{0.67} & \underline{0.21} & 0.35 & 0.55 & 0.69 \\
%  Double-scale (Vel.)                    & \underline{0.26} & \underline{0.42} & \textbf{0.58} & \textbf{0.65} & \textbf{0.20} & \textbf{0.33} & \textbf{0.53} & \textbf{0.67} \\
%  Double-scale (Hier.) (Vel.)            & 0.27 & 0.44 & 0.64 & 0.71 & \underline{0.21} & 0.34 & \underline{0.54} & \underline{0.68} \\
%  Double-scale (Phase) (Vel.)           & \textbf{0.25} & \underline{0.42} & \underline{0.60} & \underline{0.67} & \underline{0.21} & 0.34 & \underline{0.54} & 0.69 \\
 TP-RNN (Ours)                     & \textbf{\cmark} & \textbf{0.25} & \textbf{0.41} & \textbf{0.58} & \textbf{0.65} & \textbf{0.20} & \textbf{0.33} & \textbf{0.53} & \textbf{0.67} & \textbf{0.26} & \textbf{0.47} & \textbf{0.88} & \textbf{0.90} & \textbf{0.30} & \textbf{0.66} & \underline{0.96} & \textbf{1.04} \\
 \hline
\end{tabular}
}
\end{center}
\end{table*}

\begin{table*}[t!]
\caption{MAE for four action classes in the long-term forecasting experiments (prior-work results from SRNN \cite{jain2016structural}, Dropout-AutoEncoder \cite{ghosh2017learning}, and code from \cite{martinez2017human}). In each column, the best obtained results are typeset with boldface and the second best are underlined. AA: Action-Agnostic, N/A: Not Applicable, Dropout-AE: Dropout-AutoEncoder.}
\label{tab:res2}
\begin{center}
%\resizebox{\linewidth}{!}{\begin{minipage}{\linewidth}
\vspace{-15pt}
\setlength{\tabcolsep}{2.15pt}
{\small
\begin{tabular}{ l|c|ccccc|ccccc|ccccc|ccccc }
 \hline
 & AA & \multicolumn{5}{c|}{Walking} & \multicolumn{5}{c}{Eating} & \multicolumn{5}{c|}{Smoking} & \multicolumn{5}{c}{Discussion} \\
 \hline
 milliseconds              &  &   80 &  160 &  320 &  560 & 1000 &   80 &  160 &  320 &  560 & 1000    &   80 &  160 &  320 &  560 & 1000 &   80 &  160 &  320 &  560 & 1000 \\
 \hline
 ERD \cite{fragkiadaki2015recurrent}                     & \xmark   & 1.30 & 1.56 & 1.84 & 2.00 & 2.38 & 1.66 & 1.93 & 2.88 & 2.36 & 2.41 & 2.34 & 2.74 & 3.73 & 3.68 & 3.82 & 2.67 & 2.97 & 3.23 & 3.47 & 2.92\\ 
 LSTM-3LR \cite{fragkiadaki2015recurrent}                & \xmark    & 1.18 & 1.50 & 1.67 & 1.81 & 2.20 & 1.36 & 1.79 & 2.29 & 2.49 & 2.82  & 2.05 & 2.34 & 3.10 & 3.24 & 3.42 & 2.25 & 2.33 & 2.45 & 2.48 & 2.93 \\ 
 SRNN \cite{jain2016structural}                      & \xmark  & 1.08 & 1.34 & 1.60 & 1.90 & 2.13 & 1.35 & 1.71 & 2.12 & 2.28 & 2.58 & 1.90 & 2.30 & 2.90 & 3.21 & 3.23 & 1.67 & 2.03 & 2.20 & 2.39 & 2.43\\
 Dropout-AE \cite{ghosh2017learning}         & \xmark  & 1.00 & 1.11 & 1.39 & 1.55 & 1.39 & 1.31 & 1.49 & 1.86 & 1.76 & 2.01  & 0.92 & 1.03 & 1.15 & 1.38 & 1.77 & 1.11 & 1.20 & 1.38 & 1.53 & \textbf{1.73}\\ 
 Residual \cite{martinez2017human}      & \xmark   & \underline{0.32} & \underline{0.54} & \underline{0.72} & \underline{0.86} & \underline{0.96} & \underline{0.25} & \underline{0.42} & \underline{0.64} & \underline{0.94} & \underline{1.30}  & \underline{0.33} & \underline{0.60} & 1.01 & 1.23 & 1.83 & 0.34 & 0.74 & 1.04 & 1.43 & 1.75 \\
 Zero-velocity            & N/A    & 0.39 & 0.68 & 0.99 & 1.35 & 1.32 & 0.27 & 0.48 & 0.73 & 1.04 & 1.38 & \textbf{0.26} & \textbf{0.48} & \underline{0.97} & \underline{1.02} & \underline{1.69} & \underline{0.31} & \underline{0.67} & \textbf{0.94} & \underline{1.41} & 1.96\\
%  \hline
%  Single Layer (Pose)         & 0.42 & 0.70 & 0.93 & 1.06 & 1.27 & 0.32 & 0.52 & 0.74 & 1.10 & 1.40 \\
%  Single Layer (Vel.)         & 0.25 & 0.42 & \underline{0.60} & 0.76 & 0.80 & \textbf{0.20} & \underline{0.33} & \underline{0.54} & \textbf{0.84} & \underline{1.15} \\
%  \hline
%  Stacked 2-Layer (Vel.)      & \textbf{0.25} & 0.40 & \underline{0.60} & \underline{0.75} & 0.79 & \underline{0.21} & 0.35 & 0.55 & 0.86 & 1.16 \\
%  Double-scale (Vel.)         & \underline{0.26} & \underline{0.42} & \textbf{0.58} & \textbf{0.74} & \underline{0.78} & \textbf{0.20} & \textbf{0.32} & \textbf{0.53} & \textbf{0.84} & \underline{1.15} \\
%  Double-scale (Hier.) (Vel.) & 0.27 & 0.44 & 0.64 & 0.82 & 0.92 & \underline{0.21} & 0.34 & \underline{0.54} & 0.86 & 1.21 \\
%  Double-scale (Phase) (Vel.)& \textbf{0.25} & \underline{0.42} & \underline{0.60} & 0.77 & 0.82 & \underline{0.21} & 0.34 & \underline{0.54} & \underline{0.85} & \textbf{1.14} \\
 TP-RNN (Ours)         & \cmark   & \textbf{0.25} & \textbf{0.41} & \textbf{0.58} & \textbf{0.74} & \textbf{0.77} & \textbf{0.20} & \textbf{0.33} & \textbf{0.53} & \textbf{0.84} & \textbf{1.14} & \textbf{0.26} & \textbf{0.48} & \textbf{0.88} & \textbf{0.98} & \textbf{1.66} & \textbf{0.30} & \textbf{0.66} & \underline{0.98} & \textbf{1.39} & \underline{1.74}\\
\hline
\end{tabular}
}
%\end{minipage}}
\end{center}
\end{table*}

\subsection{Results}
%In the following, first the baseline and previous methods for this task are listed (used for comparisons), after which extensive experimental results show that our proposed method outperforms others in short- and long-term forecasting.

%\subsubsection{Baseline Methods}
\noindent\textbf{Baseline Methods:}
We use the following recent research to compare with: ERD \cite{fragkiadaki2015recurrent}, LSTM-3LR \cite{fragkiadaki2015recurrent}, SRNN \cite{jain2016structural}, Dropout-AutoEncoder \cite{ghosh2017learning}, 3D-PFNet \cite{chao2017forecasting}, and Residual \cite{martinez2017human}. Similar to \cite{martinez2017human}, we include the zero-velocity model as a na\"ive baseline for comparison. We also include our implementations of different LSTM-based models as part of the comparison (\ie, conducting ablation tests).

First set of our experiments compares the single layer LSTM model with pose as the input (denoted by {Single Layer (Pose)}) and the same model but with velocity as the input ({Single Layer (Vel.)}), to demonstrate that conducting the experiments in the velocity space and feeding it into LSTM sequences can better capture the human motion dynamics. As mentioned earlier, when both inputs and outputs are all velocities with similar small numerical scales, it is easier for the model to be trained. In the second set of experiments, we build multiple 2-Layer LSTM models with different architectures using velocity as the input, including the most basic one that simply stacks 2 layers of LSTMs (Stacked 2-Layer (Vel.)), commonly called multi-layer LSTM \cite{donahue2015long,wang2015long}. On top of the basic model, we build further extensions with hierarchical and multi-scale structures: two independent LSTMs (Double-scale (Vel.)), which, unlike the regular multi-layer LSTM, its higher level one does not use the output from the lower level as the input. Instead, the higher level LSTM's input is the larger scale of velocity, \ie, the velocity calculated by the pose sequence only at the odd time-steps, or only at the even time-steps. The next model (Double-scale (Hier., Vel.)) is similar to HM-RNN \cite{chung2017hierarchical}, but with slight modification of setting the higher level LSTM scale to a constant number 2, due to the fact that there is no natural boundary in human motion sequences. Another model (Double-scale (Phase, Vel.)) has multiple phases in the higher level LSTMs, capturing larger scale velocity information, rather than using the lower level LSTM outputs. Finally, we implement our proposed model (TP-RNN) with double scale setting. Note, to showcase the superiority of the proposed technique we report the results for $K=M=2$ in TP-RNN, which demonstrates that without the need to increase the network parameters, our network already outperforms all other methods. However, we also conduct an experiment for analyzing the effect of the number of levels in TP-RNN, and show models with more hierarchies can lead to even better results.

%\subsubsection{Comparison with baseline and state-of-the-art methods on the Human 3.6M dataset}
\noindent\textbf{Comparison on Human 3.6M dataset:}
Previous literature published their performance numbers on either short-term (up to 400ms) \cite{martinez2017human} or long-term (up to 1000ms) \cite{jain2016structural} predictions. Besides, some of them only report the prediction on a small set of actions (\ie, `walking', `eating', `smoking', and `discussion') \cite{jain2016structural,fragkiadaki2015recurrent}, while others report the results for all 15 actions \cite{martinez2017human} in the Human 3.6M dataset \cite{ionescu2014human3}. To compare with all the above different settings, for each of our architectures, we train a single action-agnostic model using sequence data from all of the 15 actions, without any supervision from the ground-truth action labels. We use the loss over each forecasted frame up to 1000ms (25 frames). %, the same setting as \cite{jain2016structural}. 
We follow the settings of \cite{jain2016structural,martinez2017human} for the length of the input seed observed pose (\ie, 2000ms, 50 frames).

%Here we organize our comparisons in the following tables, doing our best to cover the comparison against slight different settings from previous literatures.

Table \ref{tab:res1} shows the MAE for `walking', `eating', `smoking', and `discussion' for short-term predictions. %The performance results for the baselines come from \cite{martinez2017human}. 
Our model (TP-RNN) outperforms all the baseline results, including the current state-of-the-art, Residual model \cite{martinez2017human}, in short-term forecasting. In the Residual model \cite{martinez2017human}, pose information is used to predict the velocity of the next frame. Note that the numerical scale of velocity is much smaller compared to the pose. %, which implies that (some) trainable parameters in the model may need to have very small absolute values in order to make precise predictions. 
On the contrary, in our proposed model, velocity information of the past is fed into the models to predict the next velocity. Therefore, %our trainable parameters do not have to have small absolute values, as 
the scales of inputs and outputs are the same, which potentially puts the neural network in an easier path to train. 
%[Already moved this to table 3 discussion]In addition to the velocity extension in the single layer model, we also build the extensions with 2-Layer LSTM. Simply using stacked multi-layer LSTM does not improve the performance in general. Our other models with hierarchical and multi-scale structures have some improvements over our single layer model for the short-term prediction. The comparisons between each of our hierarchical multi-scale models is difficult to evaluate here, because our proposed models are trained using all 15 action sequence data. In this table and the next table for the long-term comparison, we report our results in order to compare with baseline models from previous research. We leave the discussion of how the hierarchical multi-scale structure improves on top of the velocity-based model in the later part of this section when we show the performance results that average all the action sequences.
For actions with large movements (like `Walking' and `Eating') our model outperforms the baselines and the state-of-the-art by a large margin. However, like other previous methods, our method has hard time to forecast `difficult-to-predict' actions (like in `Smoking' and `Discussion'). Although, our results in those activities are also superior to all other methods, they are close to the zero-velocity baseline. Our proposed TP-RNN is setting a new state-of-the-art for pose forecasting on this dataset without the need of action labels at test time. Furthermore, it conducts both short- and long-term predictions simultaneously without sacrificing the accuracy of either end. 

\begin{table*}[t!]
\caption{Long-term forecasting MAE comparison for the remaining 11 actions in Human 3.6 dataset.}% (baseline results from Residual [13], long-term experiment from the code).}
\label{tab:res3}
\vspace{-20pt}
\setlength{\tabcolsep}{2.5pt}
\begin{center}
{\footnotesize
\begin{tabular}{ l|cccccc|cccccc|cccccc|cccccc }
 \hline
 & \multicolumn{6}{c|}{Directions} & \multicolumn{6}{c|}{Greeting} & \multicolumn{6}{c|}{Talking on the phone} & \multicolumn{6}{c}{Posing} \\
 \hline
 millisec                &   80 &  160 &  320 & 400 & 560 & 1000 &   80 &  160 &  320 & 400 & 560 & 1000 &   80 &  160 &  320 & 400 & 560 & 1000 &   80 &  160 &  320 & 400 & 560 & 1000 \\
 \hline
 \cite{martinez2017human}         & 0.44 & 0.69 & 0.83 & 0.94 & 1.03 & 1.49 & 0.53 & 0.88 & 1.29 & 1.45 & \textbf{1.72} & 1.89 & 0.61 & 1.12 & 1.57 & 1.74 & 1.59 & 1.92 & 0.47 & 0.87 & 1.49 & 1.76 & 1.96 & \textbf{2.35} \\
 TP-RNN           & \textbf{0.38} & \textbf{0.59} & \textbf{0.75} & \textbf{0.83} & \textbf{0.95} & \textbf{1.38} & \textbf{0.51} & \textbf{0.86} & \textbf{1.27} & \textbf{1.44} & \textbf{1.72} & \textbf{1.81} & \textbf{0.57} & \textbf{1.08} & \textbf{1.44} & \textbf{1.59} & \textbf{1.47} & \textbf{1.68} & \textbf{0.42} & \textbf{0.76} & \textbf{1.29} & \textbf{1.54} & \textbf{1.75} & 2.47 \\
 \hline \hline
 & \multicolumn{6}{c|}{Purchases} & \multicolumn{6}{c|}{Sitting} & \multicolumn{6}{c|}{Sitting down} & \multicolumn{6}{c}{Taking photo} \\
 \hline
 millisec                &   80 &  160 &  320 & 400 &  560 & 1000 &   80 &  160 &  320 & 400 &  560 & 1000 &   80 &  160 &  320 & 400 &  560 & 1000 &   80 &  160 &  320 & 400 &  560 & 1000 \\
 \hline
 \cite{martinez2017human}         & 0.60 & 0.86 & 1.24 & 1.30 & 1.58 & \textbf{2.26} & 0.44 & 0.74 & 1.19 & 1.40 & 1.57 & 2.03 & 0.51 & 0.93 & 1.44 & 1.65 & 1.94 & 2.55 & 0.33 & 0.65 & 0.97 & 1.09 & 1.19 & 1.47 \\
 TP-RNN           & \textbf{0.59} & \textbf{0.82} & \textbf{1.12} & \textbf{1.18} & \textbf{1.52} & 2.28 & \textbf{0.41} & \textbf{0.66} & \textbf{1.07} & \textbf{1.22} & \textbf{1.35} & \textbf{1.74} & \textbf{0.41} & \textbf{0.79} & \textbf{1.13} & \textbf{1.27} & \textbf{1.47} & \textbf{1.93} & \textbf{0.26} & \textbf{0.51} & \textbf{0.80} & \textbf{0.95} & \textbf{1.08} & \textbf{1.35} \\
 \hline  \hline
 & \multicolumn{6}{c|}{Waiting} & \multicolumn{6}{c|}{Walking dog} & \multicolumn{6}{c|}{Walking together} & \multicolumn{6}{c} {Average of all 15} \\
 \hline
 millisec                &   80 &  160 &  320 & 400 &  560 & 1000 &   80 &  160 &  320 & 400 &  560 & 1000 &   80 &  160 &  320 & 400 &  560 & 1000 &   80 &  160 &  320 & 400 &  560 & 1000 \\
 \hline
 \cite{martinez2017human}         & 0.34 & 0.65 & 1.09 & \textbf{1.28} & \textbf{1.61} & \textbf{2.27} & 0.56 & 0.95 & 1.28 & 1.39 & \textbf{1.68} & \textbf{1.92} & 0.31 & 0.61 & 0.84 & 0.89 & 1.00 & 1.43 & 0.43 & 0.75 & 1.11 & 1.24 & 1.42 & 1.83  \\
 TP-RNN           & \textbf{0.30} & \textbf{0.60} & \textbf{1.09} & 1.31 & 1.71 & 2.46 & \textbf{0.53} & \textbf{0.93} & \textbf{1.24} & \textbf{1.38} & 1.73 & 1.98  & \textbf{0.23} & \textbf{0.47} & \textbf{0.67} & \textbf{0.71} & \textbf{0.78} & \textbf{1.28} & \textbf{0.37} & \textbf{0.66} & \textbf{0.99} & \textbf{1.11} & \textbf{1.30} & \textbf{1.71} \\
 \hline
\end{tabular}
      }
\end{center}
%\vspace{-18pt}
\end{table*}

Table \ref{tab:res2} shows the MAE for the same set of the four actions in the long-term prediction task. %Performance measures for the baseline models come from the SRNN \cite{jain2016structural}, and Dropout-AutoEncoder \cite{ghosh2017learning}. %% already in caption
The state-of-the-art model (\ie, Residual \cite{martinez2017human}) does not report the long-term prediction performance results, therefore we use its open-source implementation code to collect the results for long-term predictions. Note that when changing the training loss from short-term predictions to long-term predictions, this model sacrifices the prediction accuracy in the short-term time-range (less than 400ms) in order to gain the extra long-term (400ms to 1000ms) prediction ability.

The long-term prediction of the Residual model \cite{martinez2017human} still outperforms other prior works in most of the cases. Another strong previous work in long-term forecasting is the Dropout-AutoEncoder model \cite{ghosh2017learning}, which generates the best 1000ms prediction for the `Discussion' action among all other models. Similar to short-term predictions, our proposed velocity-based model outperforms all the baseline and state-of-the-art methods, except for the Dropout-AutoEncoder model in 1000ms prediction with respect to only the `Discussion' action. Note that our model conducts an action-agnostic forecasting %(\ie, unlike previous models, it trains regardless of action labels) 
and does not sacrifice the short-term or long-term predictions. Our results in comparison with other methods that are either trained for each action separately (like \cite{jain2016structural,fragkiadaki2015recurrent}) or only target short- or long-term predictions (\eg, \cite{martinez2017human,ghosh2017learning}) show better overall performance. As our models are trained using all 15 actions in Human 3.6M \cite{ionescu2014human3}, without any extra supervision from the action labels, we further evaluate the proposed method by reporting the average MAE for all time-points across all 15 action categories. In Tables \ref{tab:res3} and \ref{tab:res4}, we compare our results with the current state-of-the-art model \cite{martinez2017human}, which is the only previous research experimented on all 15 action classes. Table \ref{tab:res3} shows the long-term forecasting results of the remaining 11 action categories, not included in Table \ref{tab:res2}. As can be seen, our proposed TP-RNN model performs better than \cite{martinez2017human} in most of the action categories.

Table \ref{tab:res4} summarizes the short-term and long-term results from the current state-of-the-art model \cite{martinez2017human} (from the paper and the code, both when including action labels as inputs to the model or not), and the results from our proposed velocity-based models by showing the average MAE for all time-points across all 15 action categories. Our models outperform the Residual model, in both short-term and long-term prediction tasks. We use the average MAE metric to also conduct the ablation analysis by evaluating the difference between each of our model extensions with hierarchical and multi-scale architectures. We can see that basic multi-layer LSTM model (\ie, Stacked 2-Layer (Vel.)) does not improve the overall performance compared with single layer LSTM model (\ie, Single Layer (Vel.)). After we include the multi-scale idea in our models, we can see the improvement over the single layer model (\ie, Single Layer (Vel.)) and the basic multi-layer single-scale model (\ie, Stacked 2-Layer (Vel.)). For the long-term time-range, the model directly adapted from HM-RNN \cite{chung2017hierarchical} (\ie, Double-scale (Hier., Vel.)) does not improve the performance, because the original design of the HM-RNN model is for the natural language data with obvious boundaries in the sequence. We mitigate this limitation with our proposed TP-RNN model, which %still provides hierarchical multi-scale dynamics information, and have 
has multiple phases of LSTM on the higher levels for capturing the up-to-date larger scale temporal information. TP-RNN provides best quantitative results, as shown in Table \ref{tab:res4}.

\begin{table}[t!]
\caption{Comparison of average MAE across all 15 actions of Human 3.6 dataset with prior and baseline models (ablation study). In each column, the best results are typeset in bold and the second best are underlined. For `Residual \cite{martinez2017human}', both short-term (from paper) and long-term (from code) results are reported. AA: Action-Agnostic.}
\label{tab:res4}
\vspace{-15pt}
\setlength{\tabcolsep}{2.5pt}
\begin{center}
  \begin{adjustbox}{max width=\textwidth}
{\small
\begin{tabular}{ l|c|cccccc }
 \hline
 %milliseconds
  & AA & 80 & 160 & 320 & 400 & 560 & 1000 \\
 \hline
 Residual \cite{martinez2017human} (short)    & \xmark & \textbf{0.36} & \underline{0.67} & 1.02 & 1.15 & - & - \\
 Residual \cite{martinez2017human} (short)  & \cmark & 0.39 & 0.72 & 1.08 & 1.22 & - & - \\
 Residual \cite{martinez2017human} (long)     & \xmark & 0.43 & 0.75 & 1.11 & 1.24 & 1.42 & 1.83 \\
 Residual \cite{martinez2017human} (long)   & \cmark  & 0.42 & 0.73 & 1.09 & 1.23 & 1.42 & 1.84 \\
 Zero-velocity                       & -  & 0.40 & 0.71 & 1.07 & 1.21 & 1.42 & 1.85 \\
 \hline
 Single Layer (Pose)                 & \cmark  & 0.49 & 0.83 & 1.20 & 1.34 & 1.53 & 1.92 \\
 Single Layer (Vel.)                 & \cmark   & 0.39 & \underline{0.67} & \underline{1.00} & 1.13 & 1.32 & 1.73 \\
 \hline
 Stacked 2-Layer (Vel.)              & \cmark   & 0.38 & \textbf{0.66} & 1.01 & 1.13 & 1.32 & 1.74 \\
 Double-scale (Vel.)                 & \cmark   & \underline{0.37} & \textbf{0.66} & \textbf{0.99} & \textbf{1.11} & \textbf{1.30} & 1.73 \\
 {\footnotesize Double-scale (Hier., Vel.)}         & \cmark   & \underline{0.37} & \textbf{0.66} & \underline{1.00} & \underline{1.12} & 1.32 & 1.76 \\
 {\footnotesize Double-scale (Phase, Vel.)}         & \cmark & \underline{0.37} & \textbf{0.66} & \underline{1.00} & \underline{1.12} & \underline{1.31} & \underline{1.72} \\
 TP-RNN (Ours)                   & \cmark   & \underline{0.37} & \textbf{0.66} & \textbf{0.99} & \textbf{1.11} & \textbf{1.30} & \textbf{1.71} \\
 \hline
\end{tabular}
}
\end{adjustbox}
\end{center}
\end{table}

%NEW maybe need to change font size, Title layout width
%to make it consistent with other Table3 and Table4
\begin{table*}[t!]
%\resizebox{0.92\linewidth}{!}{\begin{minipage}{\textwidth}
%\vspace{-28pt}
\caption{Comparison with prior works on the Penn Action dataset in terms of PCK@0.05. Best results are typset in bold, and the second best are underlined. TP-RNN is tested with or without incorporating an initial pose velocity (details in the text). }% (baseline results from 3D-PFNet [2], from the paper).}
\vspace{-15pt}
\label{tab:res5}
\setlength{\tabcolsep}{4.75pt}
\begin{center}
{\small
\begin{tabular}{ l|cccccccccccccccc}
 \hline
 Future frame & 1 & 2 & 3 & 4 & 5 & 6 & 7 & 8 & 9 & 10 & 11 & 12 & 13 & 14 & 15 & 16 \\
 \hline
 Residual \cite{martinez2017human} & \underline{82.4} & 68.3 & 58.5 & 50.9 & 44.7 & 40.0 & 36.4 & 33.4 & 31.3 & 29.5 & 28.3 & 27.3 & 26.4 & 25.7 & 25.0 & 24.5  \\
 3D-PFNet \cite{chao2017forecasting}     & 79.2 & 60.0 & 49.0 & 43.9 & 41.5 & 40.3 & 39.8 & 39.7 & 40.1 & 40.5 & 41.1 & 41.6 & 42.3 & 42.9 & 43.2 & 43.3  \\
 TP-RNN w/o init vel.  & {82.3} & \underline{68.9} & \underline{61.5} &  \underline{56.9} & \underline{53.9} & \underline{51.7} & \underline{50.0} & \underline{48.5} & \underline{47.3} & \underline{46.2} & \underline{45.6} & \underline{45.0} & \underline{44.6} & \underline{44.3} & \underline{44.1} & \underline{43.9} \\
 TP-RNN w/ init vel.    & \textbf{84.5} & \textbf{72.0} & \textbf{64.8} &  \textbf{60.3} & \textbf{57.2} & \textbf{55.0} & \textbf{53.4} & \textbf{52.1} & \textbf{50.9} & \textbf{50.0} & \textbf{49.3} & \textbf{48.7} & \textbf{48.3} & \textbf{47.9} & \textbf{47.6} & \textbf{47.3} \\
 \hline 
\end{tabular}
      }
\end{center}
%      \end{minipage}}
%\vspace{-10pt}
\end{table*}

In summary, Table \ref{tab:res4} shows that the previous state-of-the-art (\ie, Residual \cite{martinez2017human}) needs to compromise between short- and long-term forecasting accuracy. However, TP-RNN model performs better, especially for the long-term forecasting (1000ms in future) without sacrificing the short-term accuracy. In terms of numbers, our long-term forecasting improvement is $\frac{1.83 - 1.71}{1.83} \propto 6.56\%$, which is not negligible. Besides, Table \ref{tab:res2} shows that for certain action categories with hierarchical multi-scale motions (e.g., `Walking'), long-term forecasting improvement is even more significant: $\frac{0.96 - 0.77}{0.96} \propto 19.79\%$. To further test the significance of the improvements, we conduct a one-tailed paired t-test between our average results and those of \cite{martinez2017human}. The $p$-value of the test equals 0.0002, which by all conventional criteria the difference between the two sets of results is considered to be extremely \textit{statistically significant}.

\begin{figure}[t!]
\centering
\begin{tabular}{cc}
\hspace{-20pt}
{\scriptsize
\begin{tikzpicture}
  \begin{axis}[width=5cm, height=3.5cm,
    symbolic x coords = {2, 3, 4, 5},
    %ymin = 0.5, ymax =1,
    %legend pos = south west,
    xlabel style={align=center},
    xlabel={Number of TP-RNN Levels\\{\small (a) short-term (400ms)}},
    ylabel={Average MAE},
    y label style={at={(axis description cs:0.15,.5)}},
  ]
  \addplot[smooth,blue,mark=*] coordinates { (2,1.113)(3,1.092)(4,1.1)(5,1.097)};
  %\legend{}
  \end{axis}
\end{tikzpicture}
}
& \hspace{-15pt}
{\scriptsize
\begin{tikzpicture}
  \begin{axis}[width=5cm, height=3.5cm,
    symbolic x coords = {2, 3, 4, 5},
    %ymin = 0.5, ymax =1,
    %legend pos = south west,
    xlabel style={align=center},
    xlabel={Number of TP-RNN Levels\\{\small (b) long-term (1000ms)}},
  ]
  \addplot[smooth,red,mark=*] coordinates { (2,1.71)(3,1.703)(4,1.689)(5,1.697)};
  %\legend{}
  \end{axis}
\end{tikzpicture}
}
\\
%     (a) & (b)
\end{tabular}
\caption{Average MAEs of short- and long-term forecasting using TP-RNN with different levels: $M\in\{2,3,4,5\}$.}
\label{fig:deeper}
\end{figure}
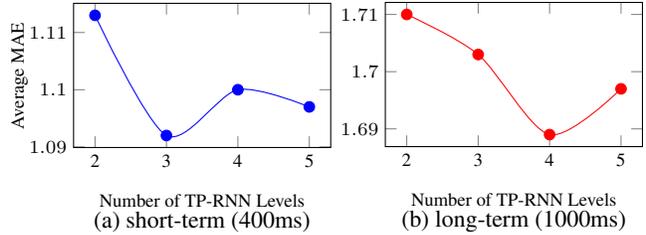

%\subsubsection{Deeper hierarchical structure of TP-RNN}
\noindent\textbf{Deeper Hierarchical Structure of TP-RNN:}
In the previous subsections, we showed that even with the most basic architectural settings (\ie, $M=2$ and $K=2$), TP-RNN already outperforms the state-of-the-art. In this section, we further experiment and analyze the effect of increasing the number of levels $M$. Fig. \ref{fig:deeper}(b) shows the average MAE of the long-term forecasting (1000ms) results from TP-RNN with different numbers of levels: $M\in\{2,3,4,5\}$. In general, increasing the number of levels of TP-RNN improves the long-term forecasting accuracy, which is in accordance with our hypothesis that higher hierarchical levels are able to better capture long-term human motion dynamics and thus improve the long-term forecasting accuracy. In addition, Fig \ref{fig:deeper}(a) shows similar results for short-term forecasting (400ms), which also indicates that the performance improves slightly when increasing the number of levels. However, when we increase the number of TP-RNN levels to 4 for short-term or 5 for long-term, we see a decline in the performance. The reason is that the LSTM cell(s) at the $5^\text{th}$ level only update once at every $2^{5-1} = 16$ time-step, which is too long, given that we only predict the future 25 frames (\ie, very few updates for $M=5$). Besides, with deeper hierarchical structures, TP-RNN has more trainable parameters and therefore is prone to overfitting (requires more data). %In this set of experiments, we started to observe overfitting when we increase the number of TP-RNN levels to above 2, so we chose to use Adam optimizer and additional dropout layers to make sure our TP-RNN model can still converge without overfitting for the cases that $M=3, M=4, M=5$.

% \begin{figure}[t!]
% \centering
% \caption{Average mean angle errors of  using TP-RNN with different levels: $M=2, M=3, M=4, M=5$.}
% \label{fig:deeper}
% \end{figure}

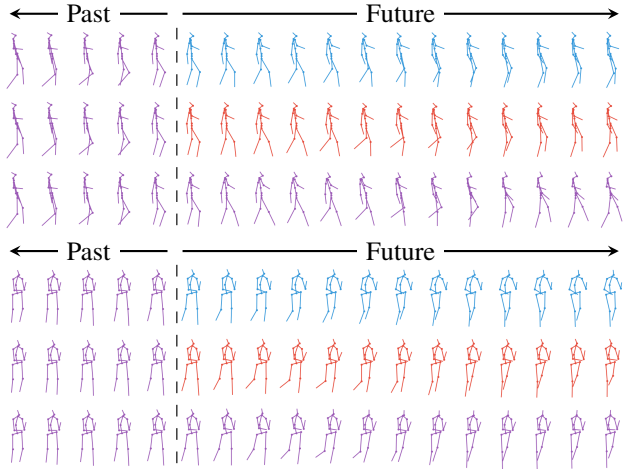
\begin{figure}[t!]
    \centering
    \begin{tikzpicture}
        \node[inner sep=0pt] (figure_walking) at (0,0)
        {\includegraphics[width=\linewidth]{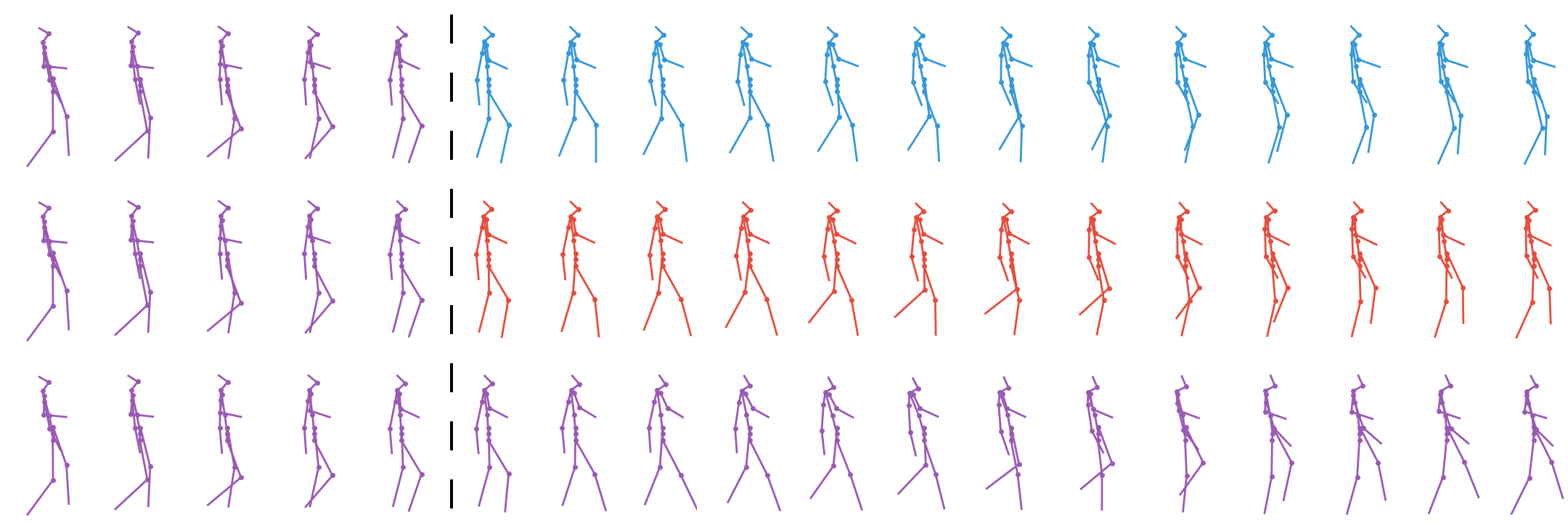}};
        \draw [<-,>=stealth,thick] (-4,1.5) -- (-1.9,1.5) node[midway,fill=white] {Past};
        \draw [->,>=stealth,thick] (-1.7,1.5) -- (4.1,1.5) node[midway,fill=white] {Future};
    \end{tikzpicture}
    \begin{tikzpicture}
        \node[inner sep=0pt] (figure_smoking) at (0,0)
        {\includegraphics[width=\linewidth]{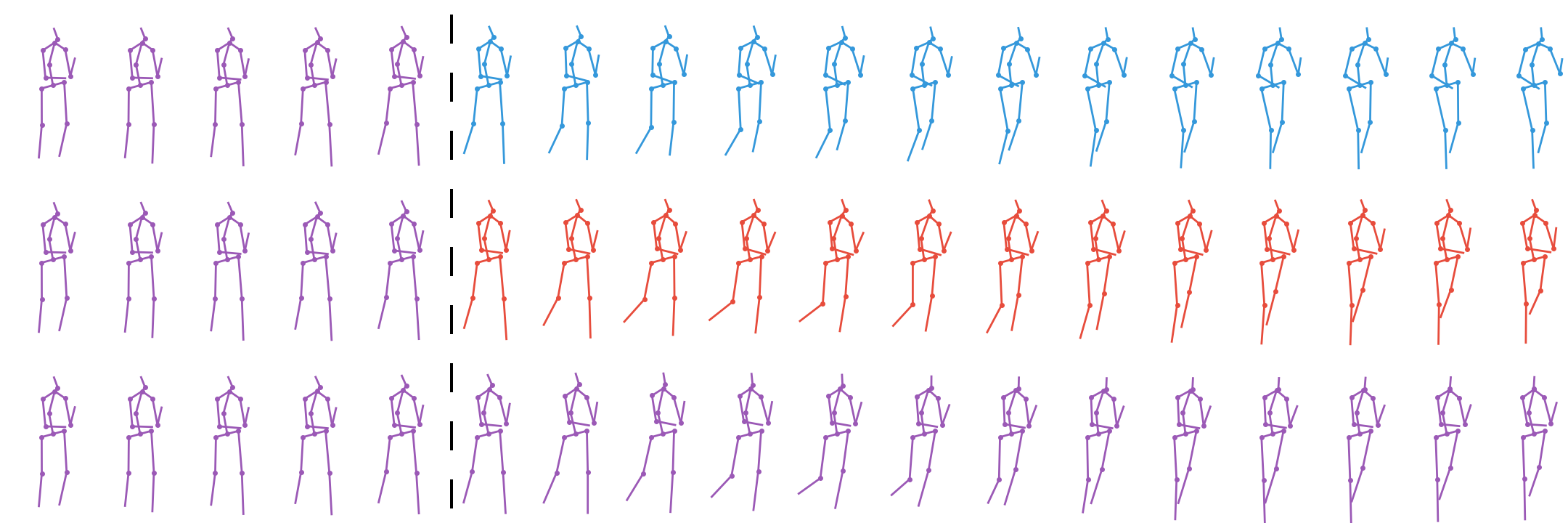}};
        \draw [<-,>=stealth,thick] (-4,1.5) -- (-1.9,1.5) node[midway,fill=white] {Past};
        \draw [->,>=stealth,thick] (-1.7,1.5) -- (4.1,1.5) node[midway,fill=white] {Future};
    \end{tikzpicture}
%    \caption{Visualization of long-term pose-forecasting (up to 1000ms) for the actions `Walking' (top) and `Smoking' (bottom), downsampled by a factor of 2 (\ie, 13 forecasted poses out of 25 are visualized). The purple poses are ground-truth data, including past (in the left side) and future time frames (in the bottom). The blue poses are the predictions of the current state-of-the-art, \ie, Residual \cite{martinez2017human}, and red poses are the results of our proposed method, TP-RNN.}
    \caption{Visualization of long-term pose-forecasting (up to 1000ms) for the actions `Walking' (top) and `Smoking' (bottom), downsampled by a factor of 2 (\ie, 13 forecasted poses out of 25 are visualized). The purple poses are ground-truth data, including past (in the left side) and future time frames (in the bottom). The blue and red poses are the predictions of \cite{martinez2017human} and our method (TP-RNN), respectively.}
    \label{fig:vis1}
\end{figure}

%\subsubsection{Qualitative evaluation and visualization}
\noindent\textbf{Qualitative Evaluations and Visualization:}
Fig. \ref{fig:vis1} shows the visualization of pose sequences for our method, in comparisons with the Residual method \cite{martinez2017human}, for the sequences for the actions `Walking' and `Smoking'. As it can be seen in the figure, our predictions (the middle row) are visually closer to the ground-truth (last row) compared to the state-of-the-art Residual method \cite{martinez2017human}. This also supports the quantitative results shown in Table \ref{tab:res2}, as our method steadily outperforms \cite{martinez2017human} for the `walking' activity: TP-RNN's MAE was 0.25 at 80ms while \cite{martinez2017human} led to an MAE of 0.32, similarly, ours was 0.75 and 0.77 in 560ms and 1000ms while \cite{martinez2017human} obtained 0.86 and 0.96, respectively. A similar conclusion can be made for predictions of the `smoking' activity. Although, this activity has slight amounts of movement, still our method can capture better dynamics, if we look at the results closely. For instance, using our method, the distance of the subject's hand from the torso and face are better predicted in long-term and leg movements are more precisely forecasted, especially in shorter-term predictions. %Another example can be seen in Fig. \ref{fig:vis3}, the long-term forecasting results of the `walking-together' action. The poses from \cite{martinez2017human} tend to converge to regular walking action poses, in which the arms are unlikely to be higher than the shoulders. However, in this `walking-together' action, the left arm is actually being placed on top of another person's shoulder and stay at a higher position. Our proposed model is capable to make better predictions in this case. 
More visualizations of the forecasted poses by TP-RNN were visualized in Fig. \ref{fig:fig1}, in which one can simply observe how the patterns of poses change through time, in comparison with the ground-truth.

\noindent\textbf{Comparison on Penn Action Dataset:}
We trained TP-RNN on Penn Action dataset \cite{zhang2013from} and, here, we compare its results with the previous state-of-the-art on this dataset, 3D-PFNet \cite{chao2017forecasting}. The input to our velocity-based TP-RNN is the pose in a single frame, and we set the initial velocity to 0 (denoted by TP-RNN w/o init vel.), in order to have a fair comparison with \cite{chao2017forecasting}. %For our TP-RNN model, the input is the pose in a single past frame, and the outputs are poses of the future 16 frames. 3D-PFNet \cite{chao2017forecasting}'s outputs are also the poses of the future 16 frames, but with slightly different setting on the input: it use the first future frame's RGB image as the input. %% Talked about these when introducing the dataset
The model performance is evaluated using PCK@0.05 as in \cite{chao2017forecasting}. PCK calculates the percentage of joint locations correctly predicted by the model. With the threshold 0.05, a joint location is counted as correctly predicted if the normalized distance between its predicted and ground-truth locations is less than 0.05. The results are shown in Table \ref{tab:res5}. %Although the input formats of our TP-RNN and 3D-PFNet are slightly different, their PCK@0.05 scores of the first future frame are close (82.3 and 79.2). %, so it is still a fair comparison to evaluate the pose forecasting capability of the two models by looking at the performance difference of all the remaining future frames. 
Our model performs significantly better than 3D-PFNet \cite{chao2017forecasting} (a $p$-value of 0.0419 < 0.05 significance threshold), with the results shown in Table \ref{tab:res5}. For further comparison, we used the open-sourced code of \cite{martinez2017human} with the necessary modifications (the same setting as our TP-RNN), and experimented on the Penn Action dataset. Additionally, we further experiment by including the initial velocity in the input, to demonstrate that the importance of the velocity information. The additional initial velocity is estimated using the difference between the current pose and the previous pose, our TP-RNN model generates even better forecasting results, as shown in the last row in Table \ref{tab:res5}. 

%With the same setting as our TP-RNN, %: using a single pose in the past frame as the input, and forecasting poses in the future 16 frames, 
%the results are shown in the first row of Table \ref{tab:res5}. %Our TP-RNN also performs better.

\section{Conclusion}
In this paper, inspired by the success of hierarchical multi-scale RNN (HM-RNN) frameworks in the natural language processing applications, we proposed a new model to encode different hierarchies in human dynamics at different time-scales. Our model trains a set of RNNs (as LSTM sequences) at each hierarchy with different time-scales. Within each level of the hierarchy, RNNs share their learnable weights, since they are all learning a same concept in a same time-scale (with different phases, \ie, different starting points of the sequence). This dramatically decreases the number of parameters to be learned in the model, while involving as much data as possible to train the higher level RNNs. As a result, the lowest layer can learn the finest grained scale motion dynamics, and higher levels capture different characteristics of the dynamics each at a certain time-scale. Furthermore, we set up a more rigorous but realistic experimental settings by conducting an action-agnostic forecasting (\ie, no usage of activity labels) and predicting both short- and long-term sequences simultaneously (unlike the previous works, which limited their settings). Despite these strict settings, our results on the Human 3.6M dataset and the Penn Action dataset show superior performance for TP-RNN to the baseline and state-of-the-art methods, in terms of both quantitative evaluations and qualitative visualizations. %For the future work, it would be interesting to explore stochastic approaches with generative models, due to the fact that human motions have varieties of patterns. 

\noindent\textbf{Acknowledgements}
We thank Panasonic for their support.

{\small
\bibliographystyle{ieee}
\bibliography{refs}
}

\end{document}